# Revisiting Active Perception [1]

Ruzena Bajcsy
*Dept. of Electrical Engineering and Computer Sciences*
*University of California, Berkeley*

Yiannis Aloimonos
*Dept. of Computer Science*
*University of Maryland*

John K. Tsotsos
*Dept. of Electrical Engineering and Computer Science*
*York University*

**Abstract**

Despite the recent successes in robotics, artificial intelligence and computer vision, a complete artificial agent necessarily must include active perception. A multitude of ideas and methods for how to accomplish this have already appeared in the past, their broader utility perhaps impeded by insufficient computational power or costly hardware. The history of these ideas, perhaps selective due to our perspectives, is presented with the goal of organizing the past literature and highlighting the seminal contributions. We argue that those contributions are as relevant today as they were decades ago and, with the state of modern computational tools, are poised to find new life in the robotic perception systems of the next decade.

## 1.0 Introduction

Some evening, long ago, our ancestors looked up at the night sky, just as they had done for thousands of years. But this time, it was different. For the first time, human eyes noticed patterns in the stars. They saw caricatures of living things and as they scanned the heavens more appeared. These were eventually named, once language permitted, and we now know them as constellations of stars. They perhaps represent the first instance of *connecting the dots*, in a literal as well as figurative manner. How did this connecting of the dots come about? What is needed to enable it? That one needs the ability to find a dot, link one dot to another is clear. Underlying this, however, is the critical ability to hypothesize a pattern: those dots look like part of something I know - let me see if I can find other dots to complete the pattern. This is the essence of active perception - to set up a goal based on some current belief about the world and to put in motion the actions that may achieve it.

Through the years, the topic of perception, and particularly vision, has been a great source of wonder and study by philosophers and scientists alike. This history cannot be reviewed here and the interested reader should see Pastore (1971) and Wade & Wade (2000), among others. This paper presents the history of the computational perspective on the problem of active perception, with an emphasis on visual perception, but broadly applicable to other sensing modalities. Those interested in a biological perspective on active perception should see Findlay and Gilchrist (2003). What follows is a brief and selective history of the birth of the active perception paradigm. However, there is one source of motivation for the topic that deserves mention. The early work of J.J. Gibson (1950) proposed that perception is due to the combination of the environment in which an agent exists and how that agent interacts with the environment. He was primarily interested in optic flow that is generated on the retina when moving through the environment (as when flying) realizing that it was the path of motion itself that enabled the perception of specific elements, while dis-enabling others. That path of motion was under the control of the agent and thus the agent chooses how it perceives its world and what is perceived within it. He coined the term *affordance*, which refers to the opportunities for action provided by a particular object or environment. These motivations play a role in an overall view of active perception, but as we will show,

---

there is more to it as well. Gibson's notions pervade our computational perspective; it is useful to note however, that Gibson in later works (Gibson 1979) became a proponent of direct perception, distinctly against an information processing view, in antithesis to what we present here.

Active Perception is a term that represents quite a broad spectrum of concepts. The SHAKEY robot, developed at Artificial Intelligence Center of Stanford Research Institute between 1966 and 1972, made history as the first general-purpose mobile robot to be able to reason about its actions (Nilsson et al. 1969). Employing cameras, range-finders and bumpers as sensors, it could be given a task and then plan how to deploy its resources, specifically in our context, its sensing resources, to complete that task. From within this team, emerged perhaps the earliest Doctoral Dissertation on active perception by J. Martin Tenenbaum in 1970, where he writes:

> *The author describes an evolving computer vision system in which the parameters of the camera are controlled by the computer. It is distinguished from conventional picture processing systems by the fact that sensor accommodation is automatic and treated as an integral part of the recognition process. Accommodation improves the reliability and efficiency of machine perception by matching the information provided by the sensor with that required by specific perceptual functions. The advantages of accommodation are demonstrated in the context of five key functions in computer vision: acquisition, contour following, verifying the presence of an expected edge, range-finding, and color recognition.*

H. Barrow and R. Popplestone (1971) also acknowledged that vision is active by writing:

> *...consider the object recognition program in its proper perspective, as a part of an integrated cognitive system. One of the simplest ways that such a system might interact with the environment is simply to shift its viewpoint, to walk round an object. In this way, more information may be gathered and ambiguities resolved. A further, more rewarding operation is to prod the object, thus measuring its range, detecting holes and concavities. Such activities involve planning, inductive generalization, and indeed, most of the capacities required by an intelligent machine.*

They did not accompany this with any strategy or method that would enable such abilities and the community let these words fade. Directions towards an embodiment would wait for quite a few more years until R. Bajcsy (1988)[2] wrote:

> *Active sensing is the problem of intelligent control strategies applied to the data acquisition process which will depend on the current state of data interpretation including recognition.*

She went on in that seminal paper to point out that this is not simply control theory. The feedback is performed not only on sensory data but on processed sensory data, and further, feedback is dependent on a priori knowledge, on models of the world in which the perceiving agent is operating. She summarizes the process nicely by saying:

> *...we have defined active perception as a problem of an intelligent data acquisition process. For that, one needs to define and measure parameters and errors from the scene which in turn can be fed back to control the data acquisition process. This is a difficult though important problem. Why? The difficulty is in the fact that many of the feedback parameters are context and scene dependent. The precise definition of these parameters depends on thorough understanding of the data acquisition devices (camera parameters, illumination and reflectance parameters), algorithms (edge detectors, region growers, 3D recovery procedures) as well as the goal of the visual processing. The*

---

[2] *An earlier version of this perspective appeared in Bajcsy (1985).*

*importance however of this understanding is that one does not spend time on processing and artificially improving imperfect data but rather on accepting imperfect, noisy data as a matter of fact and incorporating it into the overall processing strategy.*

Y. Aloimonos, I. Weiss and A. Bandyopadhyay (1988) add further structure to the concept:

*An observer is called active when engaged in some kind of activity whose purpose is to control the geometric parameters of the sensory apparatus. The purpose of the activity is to manipulate the constraints underlying the observed phenomena in order to improve the quality of the perceptual results.*

These perspectives naturally lead to a definition of active perception. Considering the wealth of insight gained from decades of research since these early papers, from many perspectives, the following emerges that will form the skeleton for this paper:

*An agent is an active perceiver if it knows why it wishes to sense, and then chooses what to perceive, and determines how, when and where to achieve that perception.*

Virtually any intelligent agent that has been developed to date satisfies at least one component of the active pentuple *why, what, how, when, where* and thus without the further connective constraint - 'and then' - this definition would not be helpful. The key distinguishing factor is the *why* component - the counterpart to hypothesizing a pattern of stars with the wish to complete it. An actively perceiving agent is one which dynamically determines the *why* of its behavior and then controls at least one of the *what, how, where* and *when* for each behavior. This explicit connection of sensing to behavior was nicely described by D. Ballard (1991) in his *animate vision* concept, writing:

*An animate vision system with the ability to control its gaze can make the execution of behaviors involving vision much simpler. Gaze control confers several advantages in the use of vision in behavioral tasks:*
1. *An animate vision system can move the cameras in order to get closer to objects, change focus, and in general use visual search*
2. *Animate vision can make programmed camera movements.*
3. *Gaze control systems can be used to focus attention or segment areas of interest in the image precategorically.*
4. *The ability to control the camera's gaze, particularly the ability to fixate targets in the world while in motion, allows a robot to choose external coordinate frames that are attached to points in the world.*
5. *The fixation point reference frame allows visuomotor control strategies that servo relative to their frame.*
6. *Gaze control leads naturally to the use of object centered coordinate systems as the basis for spatial memory.*

Ballard's points should be taken in the broader context. That is, they are not restricted to only vision but rather apply to other sensory modalities. Further, although the emphasis is on external or observable gaze, Ballard also hints at internal or non-observable components, specifically attention and choice of coordinate systems. Without these the generality seen in human visual systems cannot be achieved as argued by J. Tsotsos (2011). Table 1 presents more detail on these five basic elements, the active pentuple.

It is important to highlight that *selection* represents an integral process of all elements of the active pentuple. As Tsotsos (1980) wrote:

*Since several simultaneous [interpretation] hypotheses can co-exist, a focus of attention mechanism is necessary in order to limit the number of hypotheses under consideration.*

In addition, resource constraints play an important role not only because of computer power and memory capacity, but also because in practice, the number of sensors (and other physical components of an agent) is limited as well. Thus, choices must be made. In vision, the history of studies of visual attention covers centuries of thought and cannot be summarized here (see Tsotsos et al. 2005). Within computational vision, attention has played a role since the mid-1970's with early proponents of the explicit link between computational and biological visual attention being found in Koch & Ullman (1985), Fukushima (1986), Tsotsos (1987), and Burt (1988).

Each element of the active perception definition can be further decomposed into the set of computations and actions it comprises, as shown in Figure 1, noting that decomposition is abstract and may be further detailed. Table 2 presents each of the elements of Figure 1 along with some of the seminal works that first addressed those elements.

The remainder of this paper will present a brief historical perspective on the methods developed over the past 45+ years that address each element of what it means to be an active perceiver. The overwhelming conclusion that we draw, consistent with the seminal (Barrow & Popplestone 1971; Bajcsy 1988; Aloimonos 1990; Ballard 1991) as well as modern conceptualizations (Andreopoulos & Tsotsos 2013; Soatto 2013, among others), is that the full task of perception requires an active agent.

| Active Perception | Definition |
|---|---|
| Why | The current state of the agent determines what its next actions might be based on the expectations that its state generates. These are termed Expectation-Action tuples. This would rely on any form of inductive inference (inductive generalization, Bayesian inference, analogical reasoning, prediction, etc.) because inductive reasoning takes specific information (premises) and makes a broader generalization (conclusion) that is considered probable. The only way to know is to test the conclusion. A fixed, pre-specified, control loop is not within this definition. |
| What | Each expectation applies to a specific subset of the world that can be sensed (visual field, tactile field, etc.) and any subsequent action would be executed within that field. We may call this Scene Selection. |
| How | A variety of actions must precede the execution of a sensing or perceiving action. The agent must be placed appropriately within the sensory field (Mechanical Alignment). The sensing geometry must be set to enable the best sensing action for the agent's expectations (Sensor Alignment, including components internal to a sensor such as focus, light levels, etc.). Finally, the agent's perception mechanism must be adapted to be most receptive for interpretation of sensing results, both specific to current agent expectations as well as more general world knowledge (Priming). |
| When | An agent expectation requires Temporal Selection, that is, each expectation has a temporal component that prescribes when is it valid and with what duration. |
| Where | The sensory elements of each expectation can only be sensed from a particular viewpoint and its determination is modality specific. For example, how an agent determines a viewpoint for a visual scene differs from how it does so for a tactile surface. The specifics of the sensor and the geometry of its interaction with its domain combine to accomplish this. This will be termed the Viewpoint Selection process. |

*Table 1.* The five main constituents of an actively perceiving agent are defined.

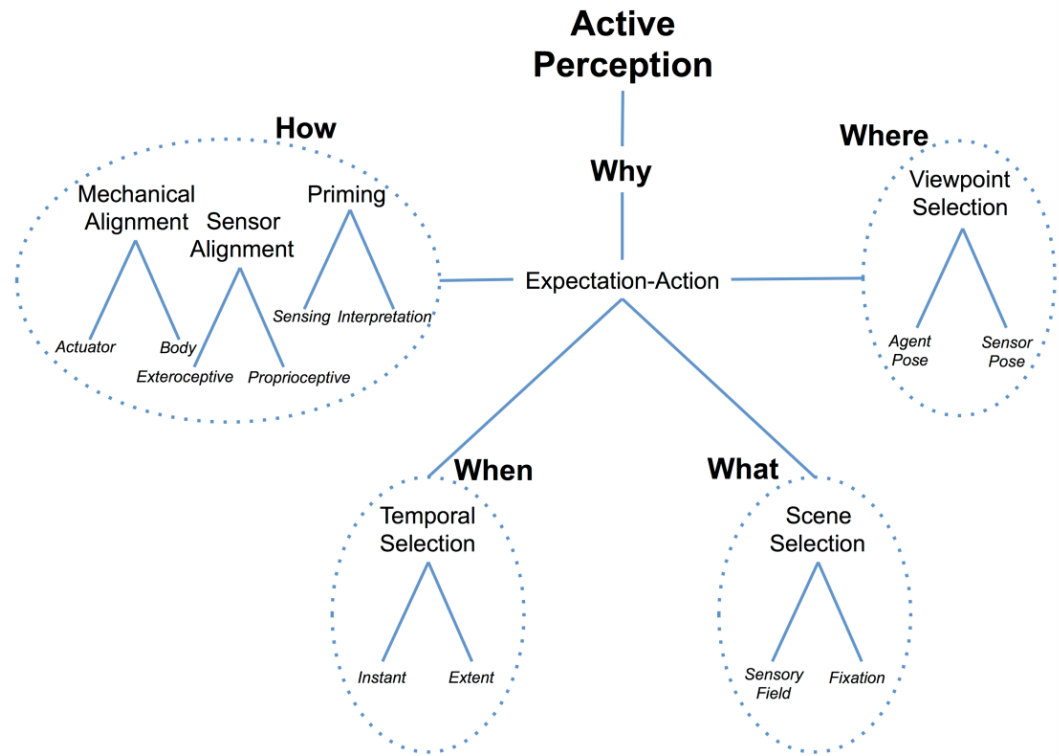

**Figure 1.** *The basic elements of Active Perception broken down into their constituent components. Instances of an embodiment of active perception would include the Why component and at least one of the remaining elements whereas a complete active agent would include at least one component from each.*

| Component | Definition | Research Seeds |
|---|---|---|
| **Mechanical Alignment** | | |
| *Actuator* | active control of motors<br>(e.g., sequence of motor actions to enable multi-view stereo) | Moravec 1980 |
| *Body* | active control of robot body and body part position and pose<br>(e.g., to move robot to a location more advantageous for current task) | Nilsson et al. 1969 |
| **Priming** | | |
| *Interpretation* | active modulation of perceptual interpretation system for current task and physical environment<br>(e.g., to tune system to be more receptive to recognition of objects and events relevant to current task) | Williams et al. 1977<br>Tsotsos 1980 |
| *Sensing* | active modulation of sensing system<br>(e.g., to tune sensors to be more sensitive to stimuli relevant to current task) | Bajcsy & Rosenthal 1975 |
| **Sensor Alignment** | | |
| *Optical Alignment* | active control of the optical elements of a visual sensor (focal length, gain, shutter speed, white balance, etc.)<br>(e.g., accommodation: increases optical power to maintain a clear image on an object as it draws near) | Tenenbaum 1970 |
| *Proprioceptive Alignment* | active control of non-contact, non-visual sensors, such as inertial measurement units<br>(e.g., the choice of path along which the IMU moves to measure linear acceleration and rotational velocity) | early 20th century, such as rocket stabilization |
| *Exteroceptive Alignment* | active control of sensors that measure the interaction with objects and environment such as applied forces/torques, friction, and shape<br>(e.g., the choice of contact pattern over time) | Allen & Bajcsy 1985 |
| **Temporal Selection** | | |
| *Instant* | active prediction of when an event is expected to occur<br>(e.g., predicting the appearance of an object in an image sequence) | Tsotsos 1980 |
| *Extent* | active prediction of how long an event is expected to occur<br>(e.g., predicting the temporal extent of movement in an image sequence ) | Tsotsos 1980 |
| **Scene Selection** | | |
| *Sensory Field* | active prediction of where in a scene a stimulus relevant to current task may appear<br>(e.g., selection of the subset of an image where a face outline can be found) | Kelly 1971 |
| *Fixation* | active prediction of which portion of a real-world scene to view<br>(e.g., indirect object search, where an easy search for a semantically related object might facilitate search for a target object) | Garvey 1976<br>Moravec 1980<br>Aloimonos et al. 1988<br>Clark & Ferrier 1988<br>Burt 1988<br>Ye & Tsotsos 1995 |
| **Viewpoint Selection** | | |
| *Agent Pose* | active selection of agent pose most appropriate for selecting a viewpoint most useful for current task<br>(e.g., moving an agent to a close enough position for viewing a task-related object or event) | Nilsson et al. 1969 |
| *Sensor Pose* | Active selection of the pose of a sensor most appropriate for the current task (includes convergent binocular camera systems)<br>(e.g., pointing a camera at a target in with the best viewing angle for its recognition) | Brown 1990<br>Coombs & Brown 1990<br>Wilkes & Tsotsos 1992 |

**Table 2.** *Details of the components of the diagram of Figure 1. For those elements where multiple Research Seeds are given, it is because each addresses a different dimension of the problem; in most cases, many open problems remain for each component.*

## 2.0 Why does an agent need to choose what to sense?

The fundamental difference between an active perception system and other perception systems lies in action, or lack of it. Whereas both types of systems include decision-making components, only the active system includes dynamic modulations to the overall agent's behavior, both external (via motors) and internal (via parameter configurations). Let us give an example: consider that we have trained, using state of the art techniques in machine learning, a filter to recognize a particular object, like a knife, from images. Consider further that our filter has a success of 90 percent. This may be a breakthrough result, however, it may be not so interesting for a behaving robotic system. Indeed, by using this filter we can search for images - in a database - containing a knife and out of ten results, nine will be correct. But this is not sufficient for an active perception-action system that needs to act and make changes to the world. With 90 percent success, 10 percent of the time the system will be acting on the wrong objects. A different approach seems needed. In the best case, this filter could be used as an attention mechanism (among others) to suggest that a knife maybe in such location. This section dissects the components of an active vision system, as shown in Figure 1, for specific cases and describes why would an agent wish to sense a particular scene or scene element. It does so by also providing some historical perspective.

As mentioned in the introduction, the essence of active perception is to set up a goal based on some current belief about the world and to put in motion the actions that may achieve it. In other words, active perception is purposive (Aloimonos 1990) as it has to combine perception and action in synthesizing the pieces that will achieve a goal. In order then to develop a theory, one would require some framework for organizing the set of goals, some form of behavioral calculus.

Early efforts to create such models gave rise to finite state machines and dynamical systems that would model agents navigating successfully in some environment. Notable among them are hybrid automata (Košecká and Bajcsy 1994) which later, in conjunction with Bayesian inference techniques gave rise to a set of techniques broadly referred to as SLAM (Simultaneous Localization and Mapping) (Thrun et al. 2000) which have enjoyed much success. Despite this success, however, the problem of localization and mapping in the absence of GPS, and when only visual sensing is available, still remains a challenge (Salas-Moreno et al. 2013), as we still lack notions of robust spatial representations. Navigation requires perception of course, but myopic vision is sufficient – or in other words, navigation needs "perception in the large". This changes, however, as one needs to interact with the environment: manipulation requires detail and so it needs "perception in the small". In the case of navigation, it was possible to develop the appropriate mathematical models. But if the agent is involved in actions where it needs to make decisions, it needs to identify, recognize and manipulate different objects, it needs to search for particular categories in the scene, it needs to recognize the effects and consequences of different actions, and all this while executing a sequence of motor actions, then it needs details from the perceptual system and it becomes challenging to manage this complexity. Where does one start? And which models to employ?

Referring back to Figure 1, we are now ready to see its components in action. Imagine an active perception agent that is at a table with the task of making a Greek salad, with all the ingredients and tools present in a cluttered scene, just like in a kitchen. The agent has a plan – or knowledge - of what is needed to produce the salad. Let's say that the plan calls first for finding a knife and cutting a cucumber. Thus, the agent would need to scan the scene in order to locate a knife. This is a top down attention problem where knowledge of a model of the object is used by the attention mechanism to find the image region likely to have a knife in it. An example of a recent approach implementing such a process can be found in (Teo et al. 2013, 2015). Thus the *What* elements of Figure 1 are engaged, described further in Section 3. But what if the agent could not find the knife? Then, it would have to move, get another view of the scene or perhaps even affect the scene by moving objects around. Such actions could uncover the knife, which happened to be behind the bowl and thus occluded. In turn, the *Where* and *How* elements of Figure 1 get engaged, further described in Sections 4 and 5. Next, after an image region containing the knife has been identified, attention brings fixation to it for the purpose of segmenting and recognizing the object. Thus, the *When* element of Figure 1 gets engaged in order to select part of the spatiotemporal data and map it to symbolic information.

The process of segmentation is as old as the field of computer vision and a large number of techniques have been developed over the years. The bulk of those techniques works, usually, with a single image and produces a complete segmentation of everything in the scene. Representatives of such algorithms can be found in (Felzenswalb & Huttenlocher 2004; Alpert et al. 2012). Algorithms of this sort perform a variety of statistical tests by comparing different parts of the image, they cannot work in real time and they need to be told in advance the number of pieces that the segmentation should contain. Clearly, such a segmentation is not appropriate for an active perception system. Ideally, one would want to segment the objects where his attention is drawn (e.g. the knife, or the knife and the cucumber in our previous example). Over the past ten years there have been a few approaches for segmenting objects driven by attention (Mishra et al. 2009, 2012a; Bjorkman and Kragic, 2010; Andreopoulos & Tsotsos 2009, 2011). Recently, the introduction of the Microsoft Kinect sensor has increased interest to the problem (Mishra et al. 2012b) because it makes it slightly easier, but challenges still remain. Even from a point cloud it may not be possible to perform segmentation without taking more visual information (or other information) into account. See for example a current approach on using the symmetry constraint for this task (Ecins et al. 2016).

For an active camera system, a summary of the many reasons for active camera control follows:
1. to move to a fixation point/plane or to track motion
   - camera saccade, camera pursuit movement, binocular vergence changes
2. to see a portion of the visual field otherwise hidden due to occlusion
   - manipulation; viewpoint change
3. to see a larger portion of the surrounding visual world
   - exploration
4. to compensate for spatial non-uniformity of a processing mechanism
   - foveation
5. to increase spatial resolution
   - sensor zoom or observer motion
6. to change depth of interest
   - stereo vergence
7. to focus
   - adjust focal length
8. to adjust depth of field
   - adjust aperture
9. to adapt to light levels
   - adjust shutter exposure time
10. to disambiguate or to eliminate degenerate views
    - induced motion (kinetic depth); lighting changes (photometric stereo) ; viewpoint change
11. to achieve a *pathognomonic*[3] *view*
    - viewpoint change
12. to complete a task
    - viewpoint change

# 3.0 How does an agent choose what to sense?

The selection of what in a sensory field to examine falls immediately into the realm of attentive processing. It has been a common tactic throughout the history of Robotics, Computer Vision and indeed all of Artificial Intelligence to deploy methods that limit processing due to limited computational resources, both memory and processing speed. Attention has been of interest to philosophers, psychologists and physiologists for a long time (for a history see Tsotsos et al. 2005). The number of experiments and theories is incredibly large and daunting in their complexity and mutual inconsistency. Nevertheless, the constant that remains is that humans, and many other animals, exhibit selective behavior in everyday life and scientists of many disciplines continue to be motivated by the search for its explanation. It should be noted that the vast bulk of research has focused on visual selective mechanisms with less on auditory attention and then far less for the other senses. Against this backdrop, several computer vision systems also included attentional strategies to limit processing in an attempt to both mimic human visual behavior as well as to economize on processing in the face of limited resources. The magnitude of the problem of resources was always felt in a practical sense yet it was only formally proved in 1989 (Tsotsos 1989) when the computational complexity of basic visual matching process was shown to be NP-Complete[4]. It was further proved there that the application of domain and task knowledge to guide or predict processing is a powerful tool for limiting processing cost turning the otherwise exponential complexity problem into a linear one.  This section will focus on visual selection.

---

[3] *Pathognomonic is a term borrowed from medicine, where it means a sign or symptom specifically characteristic or indicative of a particular disease or condition, but in our use, means a viewpoint that yields an image that is characteristic or indicative of a particular object.*

The first use in a vision system was for oriented line location in a face-recognition task (Kelly 1971). Edges in reduced resolution images were found first and then mapped onto the full image to provide face outline predictions. These guided subsequent processing. Bajcsy & Rosenthal (1975) connected visual focussing with visual attention and developed an algorithm that would focus a camera system to particular image locations, motivated by human attentive visual behavior. Garvey (1976) extended these notions to include not only locations in an image but also spatial relationships among locations, or in other words, scene context. He termed this indirect search but the general idea has broader implications. One of those implications is that the computational complexity of processing would increase without an appropriate control algorithm (the number of spatial relationships among objects in a scene is represented by an exponential function). This places further importance on selective behavior.

Perhaps the first large scale vision system was VISIONS, developed by Hanson and Riseman at the University of Massachusetts, Amherst. It incorporated image pyramid representations, attentional selection and focus of components of the hypothesis model space (Williams et al. 1977). All of this was for a single, static image. The time domain was added to this repertoire of attentional methods by Tsotsos (1980). The idea was used for temporal window prediction in a motion recognition task. Positions and poses of segments of the left ventricle during the left ventricular cycle limited the region of the X-ray image sequence to search as well as the time interval during which to search. Left ventricular motion and shape knowledge was organized hierarchically and that structure was used to generalize or specialize predictions. Bajcsy and Rosenthal built upon their earlier work, in another 1980 contribution, to tie spatial focussing to conceptual focussing in that the conceptual hierarchy was linked to spatial resolution and as a recognition process progresses, moving up and down the hierarchy, it could make request of the perceptual system for images of a particular resolution at a particular location.

Interestingly, in the early days, computational limits imposed severe constraints on what practical problems could be solved. As computers became faster and memory cheaper, those constraints shifted leading to greater range of problems that could be successfully addressed. Today's impressive successes in AI are due not in small part to such hardware advances. But the important question is: does this mean those problems can be considered as solved? The answer is 'no' and the reason what laid out by a series of papers that examined the computational complexity of basic problems in AI (see review in Tsotsos 2011). As a result, the apparent solution seen currently to problems such as object recognition is only an illusion. The problem is not solved; rather, small instances of it can be solved within some reasonable error bounds. This is a long way of saying that attentional mechanisms will have utility for a long time to come.

In computer vision, the late 1980's seemed to mark this realization. In 1987, Tsotsos showed that attention can play an important role to reduce the computational complexity of visual matching. This was quickly followed in 1988 by two works, one by Clark & Ferrier (1988) and the other by Burt (1988) who investigated ways of taking the theoretical results and using them to build functional active perception systems. Clark & Ferrier described a control system of a binocular sensor (the Harvard Head) which allows shifts in focus of attention. They are accomplished via altering of feedback gains applied to the feedback paths in the position and velocity control loops of the camera system. By altering these gains, they implemented a feature selection process motivated by the saliency maps of Koch and Ullman (1985), which was an attempt to provide a computational counterpart to the then prominent Feature Integration Theory of visual search of Treisman & Gelade (1980). This was the first realization of the saliency idea for attentive selection. Burt, on the other hand, derived a different sort of motivation from biological vision. He considered familiar aspects of eye movement control in human vision: foveation, to examine selected regions of the visual world at high resolution given a varying resolution retina; tracking, to stabilize the images of moving objects within the eye; and high-level interpretation, to anticipate where salient information will occur in a scene. The result was a high performance visual tracking system. Important and enduring elements within this work include the image pyramid and an active hypothesis-and-test mechanism.

---

[4] *Ye & Tsotsos (1996, 2001) prove that the complexity of the Sensor Planning Problem - formulated as the optimization task of selecting the set of robot actions that maximize the probability of finding a target within cost constraints - is NP-hard.*

The hypothesis-and-test idea played a role in the work of Tsotsos (1992) who examined the conditions under which image interpretation was more efficient using a passive paradigm (single image) or an active one (dynamic, interpretation guided, image sequence). He proved that even though more images are analyzed, the active approach used a *hypothesis sieve* (from an initial broadly encompassing hypothesis, evidence accumulated image by image, gradually reduces its breadth and thus amount of processing) to progressively limit processing through the image sequence so that under certain conditions, its overall complexity was far less than the blind, unguided single image case. However, active is not always more efficient than passive vision but that the constraints developed might assist an agent in choosing which to deploy. It was further shown how active perception is a subset of attentional processing in general.

The quest for methods for how to choose what to process within an image, that is, what to attend to, is now being addressed by an increasingly large number of researchers with significant progress (see the reviews in Borji & Itti 2013 or Bylinskii et al. 2015. Nevertheless, there are still many open issues (Bruce et al. 2015). In contrast, the quest for methods that determine which image to consider, that is, which visual field to sense, is not receiving much attention at all with preference seemingly given to blanket sensing of the full environment via sensors such as LIDAR. As one counterexample, Rasouli & Tsotsos (2014) show performance improvements with the integration of active visual viewpoint methods with visual attention techniques.

## 4.0 How does an agent control how sensing occurs?

A variety of actions must precede the execution of a sensing or perceiving action. The agent must be placed appropriately within the sensory field, in other words to be mechanically aligned to its task. The sensing geometry must be set to enable the best sensing action for the agent's expectations. This can be thought of sensor alignment, and encompasses the components internal to a sensor such as focus, light levels, etc. Finally, the agent's perception mechanism must be adapted or primed to be most receptive for interpretation of sensing results, both specific to current agent expectations as well as more general world knowledge.

Historically, we believe that the Stanford Hand-Eye project provided the earliest instance of a method for mechanical and sensor alignment. Tenenbaum, as part of this 1970 Ph.D. thesis at Stanford University, built an eye/head system with pan and tilt, focus control, and a neutral density filter control of a Vidicon camera. In 1973, WABOT-1 was created at the University of Waseda, which was the world's first full-scale anthropomorphic robot (Kato et al. 1973). It was able to communicate with a human in Japanese and measure the distance and direction of objects using external receptors such as artificial ears and eyes. WABOT-1 could perform tasks by vocal command that integrate sensing with actions of its hands and feet. This was followed at Stanford by Moravec's 1980 PhD work where he developed a TV-equipped robot, remotely controlled by a large computer. The Stanford Cart included a *slider*, a mechanical swivel that moved the television camera from side to side allowing multiple views to be obtained without moving the cart, thus to enabling depth computation. The cart moved in one-meter spurts separated by ten to fifteen minute pauses for image processing and route planning. In 1979, the cart successfully crossed a chair-filled room without human intervention in about five hours.

Sandini and Tagliasco (1980) demonstrated the computational benefits of a foveated camera image, borrowing the characteristics from human vision. Their model includes an explicit ability to actively scan a scene as the only way of overcoming the inherent limitations of a space-variant retina. In 1987, Krotkov as part of his Ph.D. dissertation at the University of Pennsylvania, developed the U Penn Head, with hardware similar to the previous systems, in that it had pan and tilt, focus, vergence control (Krotkov 1988, 1989). His advance to the previous work was to present solutions to two problems that arise in the context of automatically focusing a general-purpose servo-controlled video camera on manually selected targets: (i) how to best determine the focus motor position providing the sharpest focus on an object point at an unknown distance; and (ii) how to compute the distance to a sharply focused object point. Starting in 1985, Brown and Ballard led a team that designed and built the Rochester Head, a high speed binocular camera control system capable of simulating human eye movements (Soong & Brown 1991). The system was mounted on a robotic arm that allowed it to move at one meter per second in a two-meter radius workspace. This system led to an increased understanding of the role of behavior in vision, in particular that visual computations can be simpler when interacting in the 3D world. In 1995, Kuniyoshi and colleagues developed an active vision system with foveated wide-angle lenses (Kuniyoshi et al. 1995). Pahlavan and Eklund in 1992 presented the KTH-head, an anthropomorphic vision system with a focus on control of the oculomotor parameters that was inspired by human eye movements. They demonstrated an impressive, high performance, mechanical design that included 3 degrees of freedom (DOF) independently for each eye, 2 DOF for the neck and one DOF for the base. Milios, Jenkin and Tsotsos (1993) presented the system called TRISH in order to examine the robotic utility of yet another human eye movement, torsion (rotation about the optical axis of the eye). They found that control over torsion permits control over the slant of the vertical horopter in binocular processing and developed an algorithm for achieving this. There were several additional robot heads developed around the same time including those at the University of Oxford (Du et al. 1991), Harvard University (Clark & Ferrier 1988), University of Surrey (Pretlove & Parker 1993), Institut National Polytechnique de Grenoble (LIFIA) (Crowley et al. 1992), The NIST head TRICLOPS (Fiala et al. 1994), The University of Illinois head (Abbot & Ahuja 1992), and the University of Aalborg head (Christensen 1993), many of which are presented in detail in the volume of Christensen, Bowyer, and Bunke (1993). Each of these works addressed key questions regarding which oculomotor functionalities were useful to robotic perception. However, as each group learned, the engineering effort to build and maintain such systems was very large and this cost limited the proliferation of these methodologies. However, these early developments eventually made possible the development of the now ubiquitous binocular camera heads seen in most humanoid robots.

Touch can also be deployed in an active manner. From the beginning (since the 1980s) of our efforts in Active Perception we realized that vision is limited and has to be complimented by tactile perception. This has been documented by series of papers: Shape from Touch (Bajcsy 1984), Active Touch (Goldberg and Bajcsy 1984), and Feeling by Grasping (Bajcsy et al. 1984). Allen (1985) showed that tactile perception is preferred to vision when an agent must discriminate different material properties of hardness, demonstrating this on discriminating the hole on the cup handle vs. the body of the cup. Similarly, we have shown that one needs force sensing in addition to vision for exploratory mobility (Bajcsy & Sinha1989). Thus, an agent must have sufficient control mechanisms to match sensor and sensing strategy to current task. If a robot is equipped with pressure/force sensors then it can determine material properties (hard, soft, etc.) and Lederman and Klatzky (1990) proposed exploratory procedures (EP's), which associate tactile perceptions with certain motions, such as the hardness with pressing perpendicularly on the surface, tangential motion on the surface that detects surface texture, lifting that detects weight. All these capabilities to observe geometric, material mechanical (kinematics and dynamics) properties permit the determination of surface functionality. A nice example of this is Sinha (1991) who developed a hybrid position force control scheme to guarantee stability of walking. As part of this effort he had to examine the geometry of the surface as well as its material properties, such as friction and hardness of the surface. The important result of this exploratory process is that it invokes the contextual expectations that will bring constraints on geometry (for urban straight lines, for jungle clutter, etc.) or functionality of manipulation such as cutting and piercing (Bogoni & Bajcsy 1994, 1995).

# 5.0 How does an agent determine when and where to sense?

The 'when' and 'where' of our active pentuple is next in our discussion. An agent can decide from where to sense or view a scene and to do so it must first, naturally, know where it is in the context of that environment. In other words, it must know its body position and orientation with respect to the ground/gravity. This can be represented by a 6 dimensional vector (x, y, z, $\square_1$, $\square_2$, $\square_3$) determined with respect to some arbitrary location as coordinate system origin and angles with respect to the ground plane. Given the basic coordinate system, an agent needs to establish the position and viewing angles of the head/eye as well as the kinematic chain of the arms/hands. In addition, we assume that the hands have tactile sensors so that the system can establish contact with the objects and in general external world. Vision and touch, and their kinematic relationships are established through a calibration process.

Once these forms of self-information are determined, an agent can proceed to pursue the completion of its task. With respect to vision, an agent can control the viewing angle and distance for each sensing act. After the target of a sensing action, the 'what' described earlier, is set, the agent must determine in which position to place its sensors to best perceive the target and its aspects most relevant to the task at hand. The variables of interest then would be the distance to the target, $r$, and the angle of the camera optical axis, which can be represented using the polar angle $\theta$ and azimuthal angle $\varphi$. In the context of active object recognition, Wilkes and Tsotsos (1992) showed a simple behavior-based control algorithm that, depending on the state of recognition, was able to successfully control these parameters. Dickinson et al. (1994) showed the utility of aspect graph representations for encoding the appropriate active changes of sensor viewpoint for recognition. No additional variables play a role for a non-convergent binocular camera system. However, in a convergent binocular system, vergence, version, and torsion variables are also important, as shown in Milios et al. (1993), who describe the only head involving all three. There are several binocular heads that permit vergence and version control, with perhaps the KTH head of Pahlavan and Eklund (1992) being the best example.

Gaze control was addressed by Brown and colleagues (Brown 1990; Coombs & Brown 1990; Rimey and Brown 1991). The 1990 paper has particular interest because Brown details the many variables that need to be included in a control structure, using predictive methods, and considered different sorts of binocular eye movements including vergence. He contrasts several control strategies, paving the way for how the problem could be addressed in general. A solution to how to combine agent position and pose with sensing viewpoint was put forward in a series of papers by Ye and Tsotsos (1995, 1996, 1999, 2001) who began by examining the computational complexity of this sensor planning problem in the context of visual search in a 3D environment. They proved that it was NP-hard, and thus optimal solutions for all instances should not be expected. They then defined a strategy that successfully could find objects in unknown environments, a strategy refined and implemented on a variety of robot platforms (Tsotsos & Shubina 2007; Shubina & Tsotsos 2010), including the PLAYBOT intelligent wheelchair (Tsotsos et al. 1998; Andreopoulos & Tsotsos 2007) and the ASIMO humanoid (Andreopoulos et al. 2011).

As implied earlier, the current goal of an agent determines which control strategy applies best. We have addressed actions such as pick up an object, insert a piece, or decompose clutter in (Tsikos 1987; Tsikos & Bajcsy 1991). In that work we build a system composed of laser range finder (vision system) and a manipulator equipped with a selection of different grippers (two fingered hand and a suction cup) operating on a cluttered environment (postal objects, boxes, letters, tubes on a conveyor belt). The control system was modeled by a finite state machine with deterministic sensor/action connections. The goal was to have an empty conveyor belt while the system classified different objects into different pallets. The parameters to be considered were the object size in order to determine which end effector can pick them up. If the object was too big or too heavy the system had the option to push them aside. In the case that the goal is rather open ended, exploratory, then we need to equip the system at least with some broad goals to learn about its environment. This boils down to learn about the geometrical and physical properties of the environment in order to survive or act. This was executed in (Bajcsy & Campos 1992) where exploratory procedures were developed that determined material properties such as hardness, brittleness, Compliance, elasticity, plasticity, viscosity, ductility, and Impact. Geometric properties were described as shape and size, while kinematic properties were determined as degrees of freedom. The exploration control was designed as to deliver the above-mentioned attributes. For Vision task the feedback control was guided by exhaustive search, changing the views as necessary. For haptic feedback the guide was characterization of the surfaces for grasping purposes. The conclusion was that there is natural flow from active and dynamic perception, through exploration, leading to perceptual learning.

## 6.0 Empirical Issues

Computer Vision researchers have become accustomed to the use of datasets and bench-marks in their work. It is a good practical way to measure how much progress one is making. Indeed, the introduction of special datasets in stereo research or in optical flow research has acted first as a catalyst for new ideas with researchers competing for highest performance. The standard processing pipeline is shown in Figure 2.

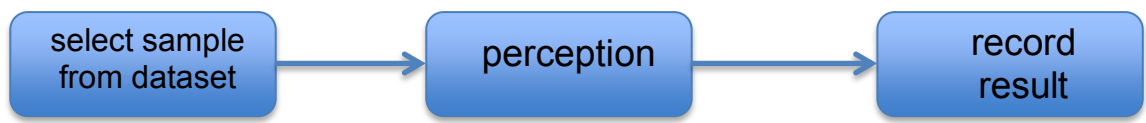

*Figure 2. The current standard processing pipeline common in computer vision.*

Having many different kinds of inputs together with ground truth information, enables almost every one with access to a personal computer to develop and experiment with new ideas. With time however, unless the datasets are replaced by new, more challenging ones, those same datasets contribute to the decline of the specific discipline. A striking example is the famous Middlebury Stereo Dataset. Researchers tailored their algorithms to perform best for the images in the dataset. When maximum performance was reached, research clearly leveled off. There was nothing else to do with this dataset, and the problem was far from solved.

Active Perception systems attempt to deal with the perceptual-motor loop of a robot in a real environment, rejecting simulated worlds. A hypothetical active processing pipeline is shown in Figure 3.

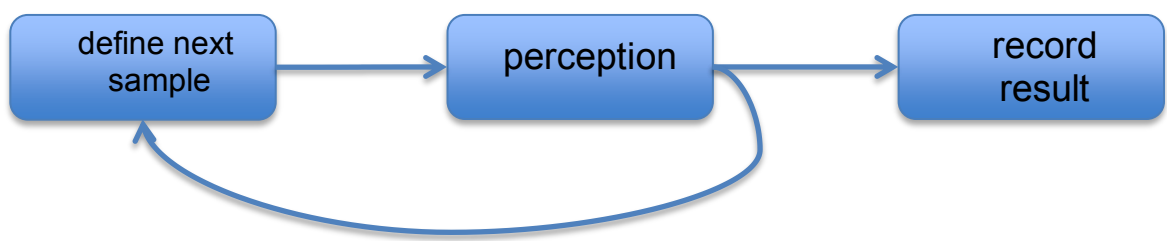

*Figure 3. Active perception processing pipeline.*

Following Gibson in his famous debate with David Marr (Bruce & Green 1990), the 'agents'-literature focuses on complex behavior coming from mechanisms operating in tight coupling with a complex environment. This is in contrast to Marr's emphasis on the feed- forward computation of a representation from sensory data. On the other hand, machine learning approaches that learn both sensory perceptions and motor actions in some environment are not easy to find. The reason is that it is difficult to build a statistical model of an environment when the system's perceptions are transformed into actions that affect the statistics of the input. Indeed, what could such an action-perception system learn? It would clearly have to develop a probability distribution of what happens. The hidden dependencies, redundancies and symmetries in this distribution are basically the structure of the world. But then, since the world is partly affected by the actions of the system, the shape of this distribution is action dependent – the system partly chooses what regularities exist, making the statistics difficult if not impossible to compute.

Active Perception by definition needs a live environment or a random scene generator that can generate on demand given imaging geometry parameters. One possibility may be provided by the Animat framework (Terzopoulos and Rabie 1997), although the lack of real environments in the perception may be problematic. In principle, an active perception system can be evaluated in terms of its performance on the tasks that it is supposed to accomplish (or the performance of its components). As such, active perception systems can be evaluated using basic techniques of systems engineering (Sztipanovits et al. 2012).

# Conclusions

Any paper whose goal is a 'revisitation' of a research topic is bound to encounter differing viewpoints and opinions that have arisen during the history of the topic. It is a challenge to appropriately include them while not detracting from the main line of the argument. Further, it is very difficult to be certain that all the relevant landmarks of research have been included; other relevant reviews and collections include Crowley, Krotkov & Brown 1992, Aloimonos 1993, Christensen et al. 1993, Crowley & Christensen 1995, Vernon 2008, Dahiya et al. 2010, Chen et al. 2011, and Andreopoulos & Tsotsos 2013, among others. Here, our main argument is that despite the recent successes in robotics, artificial intelligence and computer vision, a complete artificial agent necessarily must include active perception. The reason follows directly from the definition presented in the first section: *An agent is an active perceiver if it knows why it wishes to sense, and then chooses what to perceive, and determines how, when and where to achieve that perception.* The computational generation of intelligent behavior has been the goal of all AI, vision and robotics research since its earliest days and agents that know why they behave as they do and choose their behaviors depending on their context clearly would be embodiments of this goal.

A multitude of ideas and methods for how to accomplish this have already appeared in the past, their broader utility perhaps impeded by insufficient computational power or costly hardware. We are of the opinion that those contributions are as relevant today as they were decades ago and, with the state of modern computational tools, are poised to find new life in the robotic perception systems of the next decade. To reach this conclusion we have traversed, over a period of time extending, personally for us the co-authors, at least 4 decades, and have been seen our views strengthened. Moreover, we have been affected by a variety of motivations, disciplines, approaches, and more. These cannot all be easily presented here within a single story line, but we feel that the following personal outlooks may permit the excursions that we have individually experienced to add an interesting texture to our main story line.

***Ruzena Bajcsy***        I believe that there is a natural flow from Active Perception through Exploration to Perceptual Learning. In spite of a great deal of progress in sensing, computational power and memory, the goals set out in the original Active Perception paradigm are valid today. We are interested in conceptualizing the perceptual process of an organism/cyber-physical system that has the top-level task of functioning/surviving in an unknown environment. All the recognition algorithms depend on the context and it is context that determines the control of sensing, reasoning and action components of any system.

To conceptualize this perceptual process four necessary ingredients have emerged for either artificial or biological organisms. First, the sensory apparatus and processing of the system must be active (controllable/adjustable) and flexible.  Second, The system must have exploratory capabilities/strategies. It must also be able to evaluate at each step of exploration if it gains new information. Third, The system must be selective in its data acquisition in order to be able to act upon the perceptual information in real time. Fourth, The system must be able to learn. The learning process depends very much on the assumptions/models of what is innate and what is learned. The theory of the cyber-physical system that I have been pursuing predicts that an agent can explore and learn about its environment modulo its available sensors, kinematics and dynamic of its manipulators/end effectors, its degrees of freedom in mobility and exploratory strategies /attribute extractors. It can describe its world with an alphabet of set of perceptual and actionable primitives.

I take a great deal of inspiration from what is known about infant vision, and from psychological studies on perceptual exploration (Gibson 1950, 1979; Maturan & Varela 1987) and learning (Piaget 1962; Meltzoff and Moore 1989). Infant vision especially seems inspirational.  At birth visual structures are fully present but not fully developed. Newborns can detect changes in brightness, distinguish between stationary and kinetic objects, and follow moving objects in their visual fields. During the first 2 months, due to growth and maturation their acuity improves, they can focus and their light sensitivity improves due to pupil growth. Stereo comes into play at about 3 months and motion parallax capabilities occur around 5 month of age. Monocular depth cues come about much later, perhaps at the 4-5 month mark. The control of head movement is also gradual. Given the above facts, one can state that at best our visuo-artificial systems correspond to an infant's capabilities at about 2-5 months.  We have much to do!

Another goal is to understand how to build systems that interact with other humans. Collaboration between agents has been a challenging problem in the robotic community for many years. Perception of intent and agent affordances, the communication between agents, and the coordinated action are all open research problems. This challenge has an added layer of complexity when humans are added to the interaction the notorious *human-in-the-loop*. One of the ways we are exploring these human cyberphysical systems, is by using active perception to not only perceive the surroundings of the agents, but to examine the agents themselves. Collaborative teams of agents with different affordances offers methods of completing tasks that are far richer than homogeneous teams. By sensing the abilities of each agent, this diversity can be utilized to its furthest extent.

The affordances of the human agent are often treaded as being identical for all people. While this simplifies the modelling of the collaboration, it results in a poor tradeoff between system efficiency and safety. For instance, a device that helps an individual walk, should not impede a firefighter, but should also prevent an elderly individual from falling. While the need for this functionality is clear, the design and control of these systems requires knowledge of the individual's affordances - the system is not just human-in-the loop, but *individual*-in-the-loop.

Everyone is different, and we all change over time due to age, illness and treatment. By accounting for these changes we can better adapt a system to the interaction with the individual, while making rigorous statements about the abilities of the team, and the safety of the agents involved (Bestick et al. 2015).

*Yiannis Aloimonos*          My own views have been shaped by basic questions regarding the overall organization of an intelligent system with perception, a fundamental requirement for building intelligent autonomous robots. What kind of information should a visual system derive from the images? Should this information be expressed in some kind of internal language? Should the information be in a single general purpose form leaving it to other modules to change it to fit their needs, or can a visual system directly produce forms of information suited to other specific modules? Is part of the vision system's function to control processing in other subsystems? Is it possible to draw a sharp boundary between visual processing and other kinds of processing? How is the interaction between Vision and Cognition? The basic thesis I have been developing, along the ideas of Aristotle, Varela, Gibson and others, is that action lies at the foundation of cognition. Every component of the intelligent system is designed so that it serves the action space, which in turn serves that component (feedback loops).

The Active Vision Revolution of the '80's and 90's developed such models for the early parts of the vision system, the eye. But if the agent is involved in actions where it needs to make decisions as well as recognize and manipulate different objects, then it becomes challenging to develop the appropriate mathematical models. Where does one start? And which models to employ?

The discovery of mirror neurons (Gallese 1996) solved the dilemma, for me and others. It became clear that the system responsible for generating actions and the system responsible for interpreting actions that are visually observed, are basically the same at a high level. Thus, if it is hard to figure out how to organize the perception-action cycles of an active perception system, it may be easier to understand humans performing actions that we visually observe. In other words, if we could develop systems that understand human activity by observing it, at the same time we are developing the high level architecture of an active robotic agent that can perform the same actions. This was a powerful idea that led to a gargantuan development of computational approaches to interpreting human action from visual information and amounted to a shift of some "active vision" research to "action vision" research in my own work.

Consider again the kitchen scenario from before – the making of the Greek salad. But now consider the dual problem: an active perception system is watching a human making a salad. Its goal is to understand what it observes. Understanding has of course many layers, one needs more understanding of the action in order to replicate it than in order to name it. The active perception system watching the action would still need to engage the modules of figure 1 just as before (Yang et al. 2013). In addition, it would have to perform selection in space-time, i.e. to segment the video that it receives into meaningful chunks that can map to symbols. For example, the part of the video with the hand moving towards the knife will be segmented when the hand reaches the knife and will become the symbol REACH. The next video chunk will be GRASP. The next one will be MOVE KNIFE TO CUCUMBER'S LOCATION. Those symbols obey rules that in general form a grammar (tree like rules) or some graphical model (graphs, random fields, conditional random fields, etc.). Such knowledge allows a linguistic treatment of action, since one would be able using such models to create a parsing of the scene and a subsequent semantic analysis (Woergotter et al. 2012; Summers-stay et al. 2013; Pastra and Aloimonos 2012; Manikonda et al. 1999; Dantam and Stillman 2013; Fainekos et al. 2005; Aksoy et al. 2011; Worch et al. 2015; Yang et al. 2015; Coates et al. 2008; Maitin-Shepard et al. 2010).

Considering then an active perception system that is able to perform thousands of actions involving thousands of objects, how would we call the actions and objects involved? Would we name them action-1, action-2, .., action-m and similarly object-1, object-2, and so on? Or would we use their actual names, such as grasp, cut, move, screw, turn, press, …, cucumber, tomato, knife, and so on? The second choice allows the possibility of structuring the knowledge of an active perception system in terms of language, allowing for better communication with humans as well as for using natural language processing techniques in the integration of the system. This avenue contributes to the grounding of the meaning of language and is an important research direction. (Teo et al. 2012; Kollar et al. 2010; Tellex et al. 2009; Siskind, 2001; Yu et al. 2011; Yang et al. 2014; Zampogiannis et al. 2014; Yang et al. 2015).

***John Tsotsos***      My early years in the field, when we collectively believed that we could use computation to understand human visual and cognitive abilities, have played a large role in shaping my viewpoint today. Over the years I find myself more and more influenced by our ever-growing knowledge about the human visual system and human behavior. Perhaps more is known about the visual system of the brain than any other component, and more of the brain's cortical neurons are devoted to visual processing than any other task. Each discovery about human vision can be regarded as a hint or clue that might be helpful in developing a functioning artificial vision system. But there are far more experimental discoveries than useful clues. And how one translates these hints into real systems is also very important. At what level of abstraction should it apply? What mathematical formulation or computational construct best models the hint? How is the result evaluated with respect to its computational performance? What degree of faithfulness to those neurobiological and behavioral observations is most useful? It is important to understand that not all experimental observations about vision are useful and generally we do not have a way of knowing which. Any choice of neurobiological or behavioral hints to use in an artificial vision system constrains the space of possible systems that can result. A large aspect of the art of creating artificial vision is to select the right subset and to determine the best way to translate those hints into enabling elements.

The hallmark of human vision is its generality. The same brain and same visual system allow one to play tennis, drive a car, perform surgery, view photo albums, read a book, gaze into your loved one's eyes, go online shopping, solve 1000-piece jigsaw puzzles, find your lost keys, chase after your young daughter when she appears in danger, and so much more. The reality is that incredible as the AI successes so far have been, it is humbling to acknowledge how far there is still to go. The successes have all been uni-taskers (they have a single function) — the human visual system is a multi-tasker, and the tasks one can teach that system seem unbounded. And it is an infeasible solution to simply create a brain that includes a large set of uni-taskers. So how to move forward? We need to consider a broader set of vision problems. We can slowly move towards that goal by remembering that people — and their visual systems — move. Thus, the constraints that can be applied to vision system development can be expanded by including the constraints that self-motion imposes, such as the spatial, as well as temporal, correspondence between successive images; there is a cost in time and energy involved in moving the eyes (or body), and this constrains how often one may be willing to do so; there must be an innate understanding of visual perspective and geometry in order to build an internal representation of what we have seen in previous views and where it is; and more. If we wish to fulfill the dream of humanoid robotic companions for our elderly and infirm, or household assistants, such constraints are central.

But with this possibility a new problem arises: the amount of sensory data to be processed grows rapidly and computational power begins to be strained. My solution is to seek a general purpose vision system that can be tuned to different functions depending on the task required of it and input it views (Tsotsos 2011). One aspect of this tuning is to select which portion of the input to process at any time (select the visual field, from which viewpoint it is sensed, etc.). The brain can also select the manner in which this input is processed at any time, by priming the system for its current expectations, suppressing irrelevant computations thus enhancing the relevant ones, improving responses to task-relevant image characteristics, sharpening decision processes, and more. This is the essence of attentional behavior and active perception represents an important subset of the full range of attentional behaviors observed in humans (Tsotsos 1992). Thus, I arrive my current perspective on the problem. It's all about control! A passive sensing strategy, no matter how much data is collected, gives up control over the quality and specific characteristics of what is sensed and at what time and for which purpose. Passive sensing reduces or eliminates the utility of any form of predictive reasoning strategy (hypothesize-and-test, verification of inductive inferences including Bayesian, etc.). And it's too early in the robot intelligence story to believe such reasoning strategies are ultimately unnecessary.